\documentclass[preprint]{elsarticle}
 \pdfoutput=1
\usepackage{subfigure}
\usepackage{xcolor}
\usepackage{colortbl,booktabs}
\usepackage{amsmath}
\usepackage{tabularx}
\usepackage{array}
\usepackage{booktabs}
\usepackage{amssymb}
\usepackage{graphicx}
\usepackage{multirow}
\usepackage{footnote}

\usepackage{indentfirst}
\usepackage{lineno}
\usepackage{algorithm,algorithmic}
\usepackage{float}
\usepackage{listings}

\biboptions{numbers,sort&compress}

\newcommand{\upcite}[1]{\textsuperscript{\textsuperscript{\cite{#1}}}}

\journal{Construction and Building Materials}

\begin{document}

\begin{frontmatter}


\title{Defect segmentation: Mapping tunnel lining internal defects with ground penetrating radar data using a convolutional neural network}

\author[mythridaddress]{Senlin Yang}
\ead{yangsenlin@mail.sdu.edu.cn}

\author[mymainaddress]{Zhengfang Wang}
\ead{wangzhengfangsdu@hotmail.com}

\author[mymainaddress]{Jing Wang\corref{mycorrespondingauthor}}
\cortext[mycorrespondingauthor]{Corresponding author}
\ead{wangjingkz@sdu.edu.cn }

\author[myfourthaddress]{Anthony G. Cohn}
\ead{ A.G.Cohn@leeds.ac.uk}

\author[mymainaddress]{Jiaqi Zhang}
\ead{201934530@mail.sdu.edu.cn}

\author[mythridaddress]{Peng Jiang}
\ead{sdujump@gmail.com}

\author[mysecondaryaddress]{Lichao Nie}
\ead{llichaonie@163.com}

\author[mymainaddress]{Qingmei Sui}
\ead{qmsui@sdu.edu.cn}

\address[mymainaddress]{School of Control Science and Engineering, Shandong University, Jinan, 250061, China.}
\address[mysecondaryaddress]{Geotechnical and Structural Engineering Research Center, Shandong University, Jinan, 250061, China}
\address[mythridaddress]{School of Qilu Transportation, Shandong University, Jinan, 250061, China}
\address[myfourthaddress]{School of Computing, University of Leeds, Leeds, LS29JT, UK}


\begin{abstract}
This research proposes a Ground Penetrating Radar (GPR) data processing method for non-destructive detection of tunnel lining internal defects, called \emph{defect segmentation}. To perform this critical step of automatic tunnel lining detection, the method uses a CNN called Segnet combined with the Lovász softmax loss function to map the internal defect structure with GPR synthetic data, which improves the accuracy, automation and efficiency of defects detection. The novel method we present overcomes several difficulties of traditional GPR data interpretation as demonstrated by an evaluation on both synthetic and real datas -- to verify the method on real data, a test model containing a known defect was designed and built and GPR data was obtained and analyzed.
\end{abstract}



\end{frontmatter}


\section{Introduction}

Tunnels are an essential part of traffic and water conservancy projects and, their safe operation has always been a critical engineering issue\upcite{ richards1998inspection}. Tunnels require maintenance service due to ageing, geological conditions, and natural weathering. During the maintenance process, a variety of defects in the tunnel lining often present themselves. This leads to tunnel instability and can affect tunnel operation safety. Common types of tunnel lining internal defects include cracks, voids, lining-rock separation, water seepage, and other structural defects, which affect the stress and erosion on the tunnel differently \upcite{on1991report, Kolymbas2007Groundwater, meguid2009effect}. Effectively knowing the classification, location, and shape of any lining internal defects provides an essential basis for timely solutions and helps ensures the safety of the tunnel.

Common detection methods for tunnel lining defects include direct methods for extracting centroid detection, non-destructive testing (NDT) techniques such as infrared thermography, multi-spectral analysis, ultrasonic pulses, and Ground Penetrating Radar (GPR), etc\upcite{ popovics1990behaviour, le2002overview}. Among them, GPR has become a preferred method for defect detection in tunnel linings due to its advantages of fast detection speeds, strong penetration ability, convenient use, and carrying ease \upcite{davis2005rapid}. Research on tunnel integrity detection using GPR can be traced back to 1994\upcite{holub1994detection}, and many researchers have studied the performance of GPR\upcite{Kuloglu2010Ground, zhang2010application, alani2018gpr}, which has gradually formed a discipline.

For GPR, tunnel lining internal structures can be deduced, due to different relative dielectric constants, by emitting electromagnetic waves and receiving the reflected signals. These reflected signals are hyperbolic, and they are often interlaced with each other, which makes data interpretation difficult. Commonly used theoretical methods are migration imaging and inversion calculation, which can obtain the relative dielectric constant model. Researchers have conducted comprehensive work on this\upcite{jazayeri2019reinforced, zhang2019application}. In addition, much research exists regarding automatic identification of anomalous objects in GPR data based on the pattern recognition and machine learning methods. Pasolli et al.\upcite{pasolli2009automatic} used Genetic Algorithm and Support Vector Machine (SVM) to perform pattern recognition and classification on pre-processed GPR data and achieved relatively accurate identification. Xie et al.\upcite{xie2013gpr} used SVM to extract the void signal from synthetic GPR data, and collected real data through model tests to apply the method. Although 97.74\% accuracy was obtained, it is difficult to use their method to accurately obtain the position and shape of the voids. Dou et al.\upcite{dou2016real} and Zhou et al.\upcite{zhou2018automatic} respectively proposed a C3 clustering algorithm and an Optimized Stable Clustering Algorithm (OSCA) to extract complex GPR reflection signal characteristics and then fit them to GPR reflection hyperbola parameters.

With the rapid development of artificial intelligence and deep learning in recent years, deep learning algorithms based on Convolutional Neural Networks (CNNs) have provided new solutions for GPR data processing and defect recognition. In the field of image and computer vision, methods such as FCN (Fully Convolutional Network)\upcite{long2015fully}, U-net\upcite{ronneberger2015u}, and Segnet\upcite{badrinarayanan2017segnet} have led to better and better achievements and have gradually been applied to autopilot systems and other applications. CNNs have also been introduced in the medical field to perform defect detection and identification\upcite{ronneberger2015u, xu2014deep, kayalibay2017cnn}. Likewise, in geophysics, there have been many studies to solve the inversion problem using CNNs and related methods\upcite{LiDeep, liu2019deep, liu2019gprinvnet}. For GPR data recognition, Nuaimy et al.\upcite{al2000automatic} effectively combined GPR data processing, pattern recognition, and neural networks to complete high-resolution labelling, imaging, and classification of GPR data as early as 2000. It provided a reference for the application of neural networks in solving the GPR data interpretation problem. Xu et al.\upcite{xu2018railway} used vehicle-borne GPR to detect railway subgrade defects and apply the Faster R-CNN method to identify defect signals in GPR data. Their research effectively obtained the position, classification, and probability of defects in GPR data image. In terms of GPR detection on asphalt pavements, Tong et al.\upcite{tong2017recognition} designed a recognition CNN, a location CNN, and a feature extraction CNN, to solve the automatic recognition, location, length measurement, and 3D reconstruction of cracks, respectively. 

The research methods reviewed above mainly identified defect signals in GPR data images allowing for accurate classification and positioning. It is necessary to improve the accuracy, automation, and efficiency of GPR data interpretation for tunnel linings internal defects, and to obtain the classification and structures of linings internal.

Inspired by the progress in semantic segmentation in computer vision, we considered the possibility of mapping the tunnel lining internal structure, including the classification, location and shape of the defects, with GPR data. Therefore, this paper proposes an innovative method, using a CNN to complete GPR data processing and obtain pixel-level tunnel lining internal defects information, which we called defect segmentation.
In this way, after GPR data is routinely processed and used as input, detailed information on the tunnel lining internal material structure can be obtained, which is more automated and intuitive.
 Through our method, the efficiency and enforceability of tunnel defect detection can be greatly improved. The main work we report on here includes effective data preparation, selection and analysis of CNNs, and application to real data. The rest of the paper is arranged as follows: In Section ~\ref{section: method}, we describe the characteristics of our proposed method, and provide designed dielectric constant models and corresponding synthetic GPR data for training of CNNs. Then we design CNNs based on the characteristics of GPR data and introduced the detailed parameters of the CNNs in Section ~\ref{section: CNN}. The performance of the proposed CNN is discussed, comparing different CNNs and different defects. Section ~\ref{section: results} reports on the results obtained from analyzing synthetic data whilst Section ~\ref{section: model} focuses on evaluating performance of the method on real data derived from a test model. Finally, in the conclusion was summarizedwe summarize the contributions of the paper.

\section{Method description and GPR data preparation}
\label{section: method}
\subsection{Method description}

\begin{figure*}[hb]
	\centering
	\includegraphics[width=0.8\linewidth]{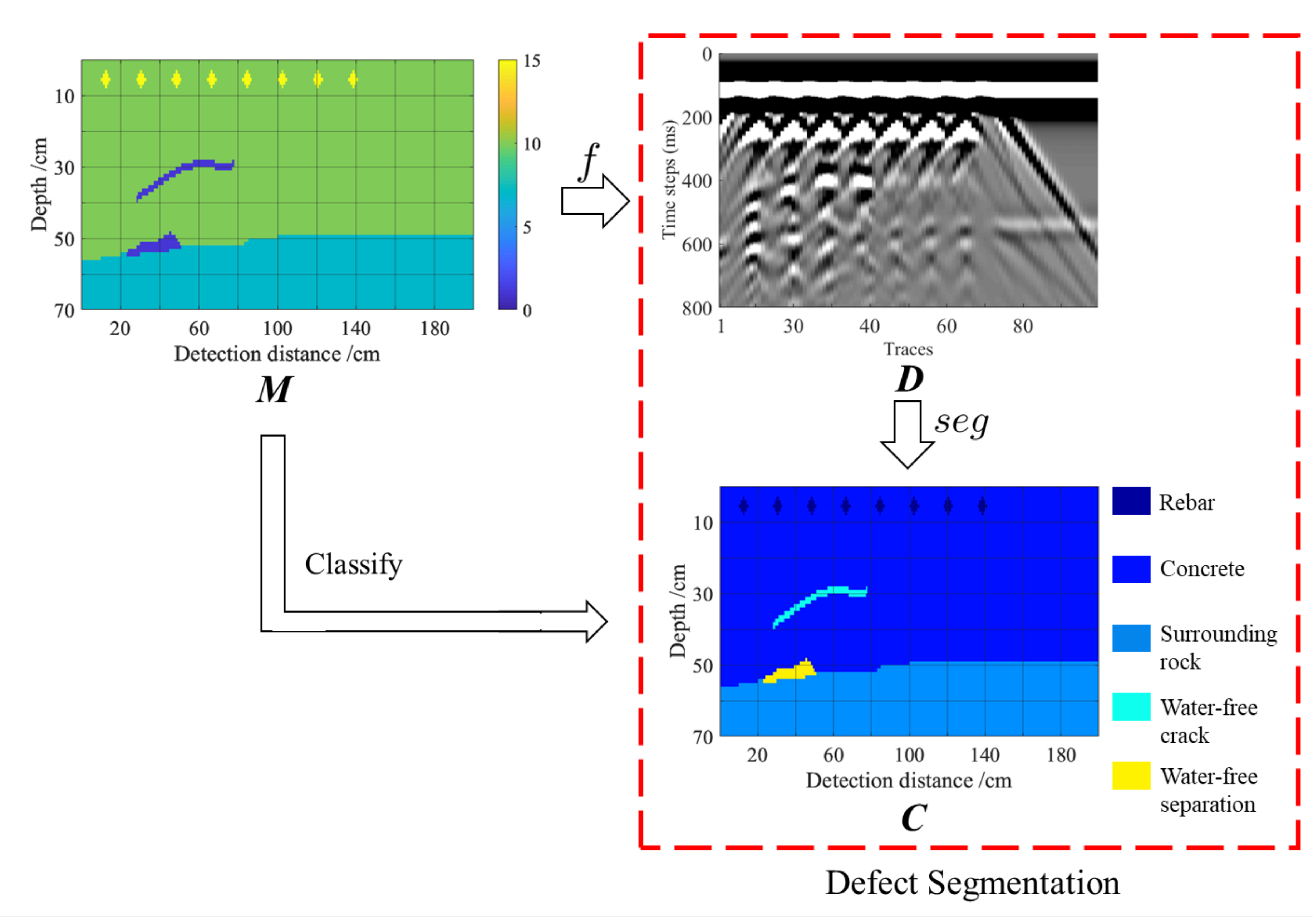}
	\caption{Defect segmentation method. The numerical model $M$ was designed, and the GPR data $D$ was obtained by modelling. At the same time, according to the model $M$, the defect segmentation model $C$ was obtained and used as a label for training the CNN. Our defect segmentation method trains the CNN to get the mapping relationship $seg$ and complete the calculation of $C$ from $D$.}
	\label{fig:1}
\end{figure*}

The aim of this research was to use GPR equipment to collect reflection data from the internal structures of the tunnel lining and then to deduce the internal materials or defects to a complete pixel-by-pixel level defect segmentation. In general, the results of GPR detection are different due to the differing internal materials and defect shapes. For complicated internal structures, they are stacked on top of each other and have similar shapes, which makes it difficult to separate them effectively and give the correct explanation. The defect segmentation task we propose is aimed at this complex problem that may include rebars, surrounding rocks, and multiple defects. Let us denote the relative dielectric constant model inside a tunnel lining can be described as $M$; the resulting GPR we denote as $D$; $D$ can be segmented to provide a classification, $C$, of the image data reflecting the elements of the model. Thus we can describe the problem as: 
\begin{equation}
D=f(M),
\end{equation}

\begin{equation}
C=seg(D),\\
\end{equation}
where, $seg$ is the mapping from GPR data to defect segmentation, and $f$ represents the process of collecting GPR data for the internal model of the lining, which is shown in Fig.~\ref{fig:1}. 

\begin{figure*}[hb]
	\centering
	\includegraphics[width=0.5\linewidth]{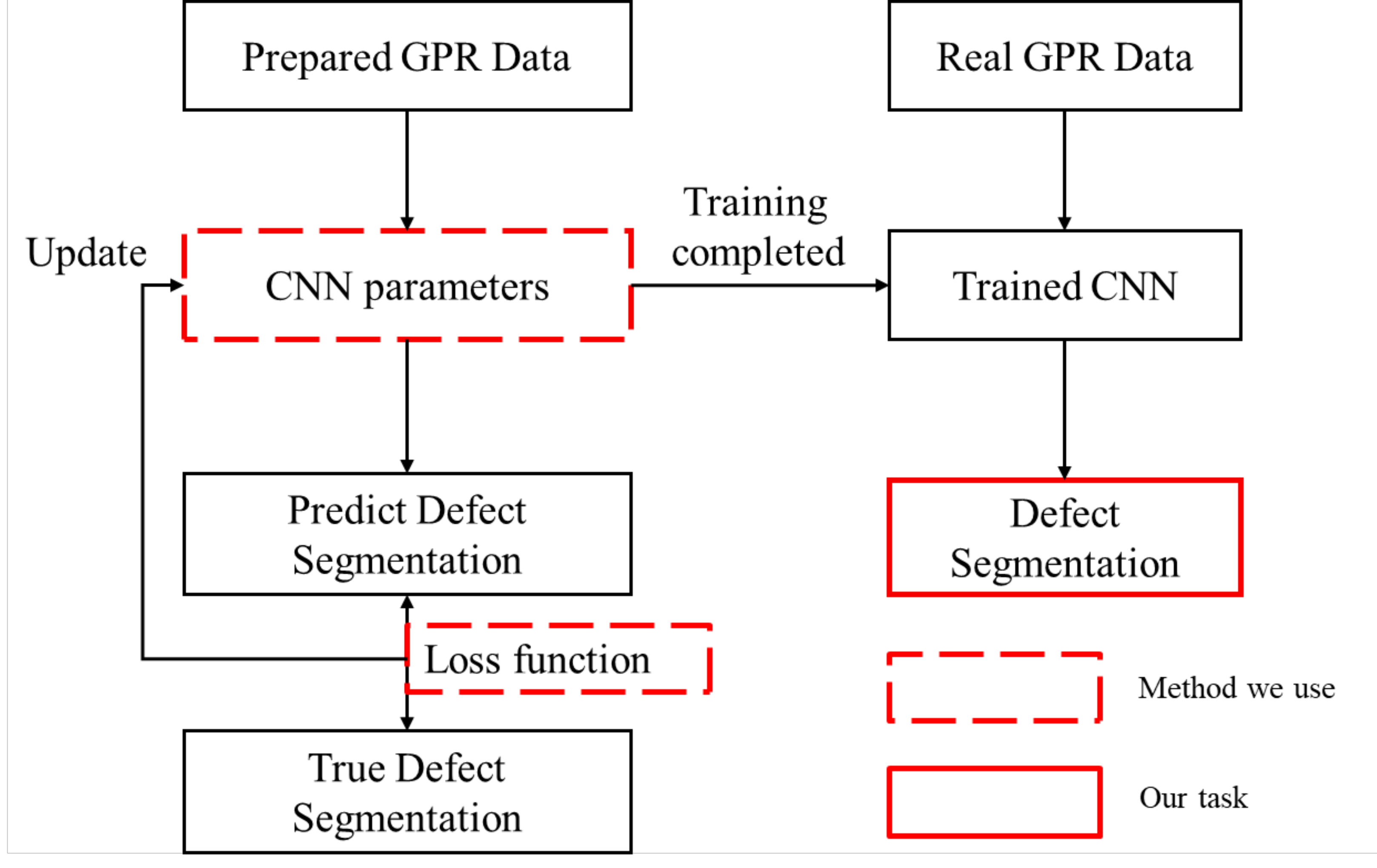}
	\caption{The workflow of the CNN based segmentation method. Prepared GPR data is input into the CNN, and the predicted result, based on current CNN parameters, is obtained. The difference between the prediction result and the actual model is calculated by the loss function, and the CNN parameters are updated by the gradient. The CNN is trained after multiple iterations. After inputting the GPR data, the CNN parameters can be used to directly obtain the fault segmentation.}
	\label{fig:2}
\end{figure*}

 CNNs based on supervised learning need to train network parameters from many model-data pairs to obtain a network model and complete $seg$, i.e., the mapping of $D$ to the classification model $C$. After that, the real data $D$ was only needed to be used as an input to the trained CNNs; the correct defect category, location, and shape can be quickly obtained. This makes the interpretation of radar data simple, automatic and efficient. The workflow is shown in Fig.\ref{fig:2}.
 
 For classification problems it is worth noting that because the dielectric constant of the same material is within a specific range, it is a many-to-one problem, that is, for the same $C$, $D$ is not fixed, which is different from the inversion problem. 


\subsection{Tunnel lining interior materials and defects}
\label{section:data set}
Because the training of CNNs relies on a large amount of data, and the structure inside the tunnel lining is difficult to obtain directly, we usually collect models and data based on numerical simulation methods. The prediction result of CNNs depends on the model used to train it, so the practical analysis and selection of tunnel lining materials and the correct design of the tunnel lining models are also the focus of our research.

Common tunnel lining interior materials and defect types include rebars, rock, voids, cracks, lining-rock separation, and water seepage. They can be summarized as five types of media, namely air, water, concrete, surrounding rock, and rebar. Their relative dielectric constants and conductivity ranges, which are modified based on \cite{davis1989ground}, are shown in the Table ~\ref{tab:table1}. Among them, water and air are defective media, and they are contained in voids, cracks and lining-rock separations. The interior of the tunnel lining may also contain rebars and rocks, which affects the GPR data and make it more complicated. Dissimilar materials have different effects on GPR due to their different dielectric constants and electrical conductivity, which is also the basis for our defect identification.

\begin{table}[]
\centering
\caption{Relative dielectric constant and conductivity properties of different media}
\label{Tab:01}
\begin{tabular}{cccll}
\cline{1-3}
Media            & Relative dielectric constant & Conductivity S/m      &  &  \\ \cline{1-3}
Air              & 1                            & 0                     &  &  \\
Water            & 81                           & 0.0005                &  &  \\
Rebar            & 300                          & 10\textasciicircum{}8 &  &  \\
Surrounding rock & 6$\sim$8                     & 0.001                 &  &  \\
Concrete         & 8$\sim$10                    & 0.0001                &  &  \\ \cline{1-3}
\end{tabular}
\label{tab:table1}
\end{table}

The materials and defects in the designed model can be divided into nine types: rebar, concrete, rock, crack, water-bearing crack, void, water-bearing void, lining-rock separation, and water-bearing separation. The lining-rock separation is a defect that appears between the lining and the surrounding rock, and a void is inside the concrete. They cause different hazards, so although they are similar in shape, different types are used here\upcite{meguid2009effect, gao2014estimation}. The crack is a small crack that arises from the uneven forces in the tunnel. It also appears in the concrete and seriously affects the tunnel lining bearing capacity\upcite{yu2016efficient}. The above three types are further expanded into six types of defects according to whether they contain water or not. According to these materials and defect types, they are divided into the following categories:

\begin{itemize}
\item No defect in tunnel lining;
\item A water-free defect in tunnel lining without rebar;
\item A water-bearing defect in tunnel lining without rebar;
\item A water-free defect in tunnel lining with rebar;
\item A water-bearing defect in tunnel lining with rebar.
\end{itemize}

Based on the categories summarized above, we designed 124,800 sets of tunnel lining relative dielectric constant models, as training models for deep learning algorithms, which covers most cases. The grid size of the model was $70 \times 200$, and the length and width of each grid were 0.01$m$, that is, the model had an actual width of $2.0m$ and depth of $0.7m$. In addition, considering the need to use a Convolutional Perfect Matching Layer (CPML) to reduce false boundary reflections, the model was expanded by ten additional meshes in speculation, and the final model size was $90 \times 220$.

 The excitation end of the GPR emits high-frequency electromagnetic waves, in the form of pulses, and propagates in the medium. In the process of propagation, and due to the difference in electromagnetic characteristics of dissimilar materials (such as air, water), the propagation of the electromagnetic waves is affected, and reflected waves are generated. This reflected signal is then received by the GPR antenna. For the CNN training, the closer the simulated data is to the real GPR data, the stronger the applicability of the network. This paper uses the Finite Difference
Time Domain (FDTD) method to calculate the GPR data of the dielectric constant model in Section ~\ref{section: results in u}. For models of size $90 \times 220$, a Ricker wavelet with a main frequency of 600 MHz is used, and each model uses 99 sidelines. The sampling time interval is 2.3587$ \times 10^{-11}$, with a total sampling of 800 steps.

\section{Convolutional Neural Networks}
\label{section: CNN}
CNNs have become one of the most significant application methods of deep learning due to its wide use in image processing. CNNs can extract features from the input data and then perform tasks such as classification, recognition or prediction. CNNs have been proposed for semantic segmentation. They have achieved impressive results in the natural and medical images field, and they have also been gradually used in the field of geophysics. Among them, CNNs include FCN\upcite{long2015fully}, U-net\upcite{ronneberger2015u} and Segnet\upcite{badrinarayanan2017segnet}. In this study, we analyzed the defect segmentation method, introduced Segnet and adopted a new loss function to obtain more accurate results. Finally, hyperparameters were set to update the network parameters reasonably and effectively in the training process.

\subsection{CNNs for defect segmentation}

\begin{figure*}[hb]
	\centering
	\includegraphics[width=1\linewidth]{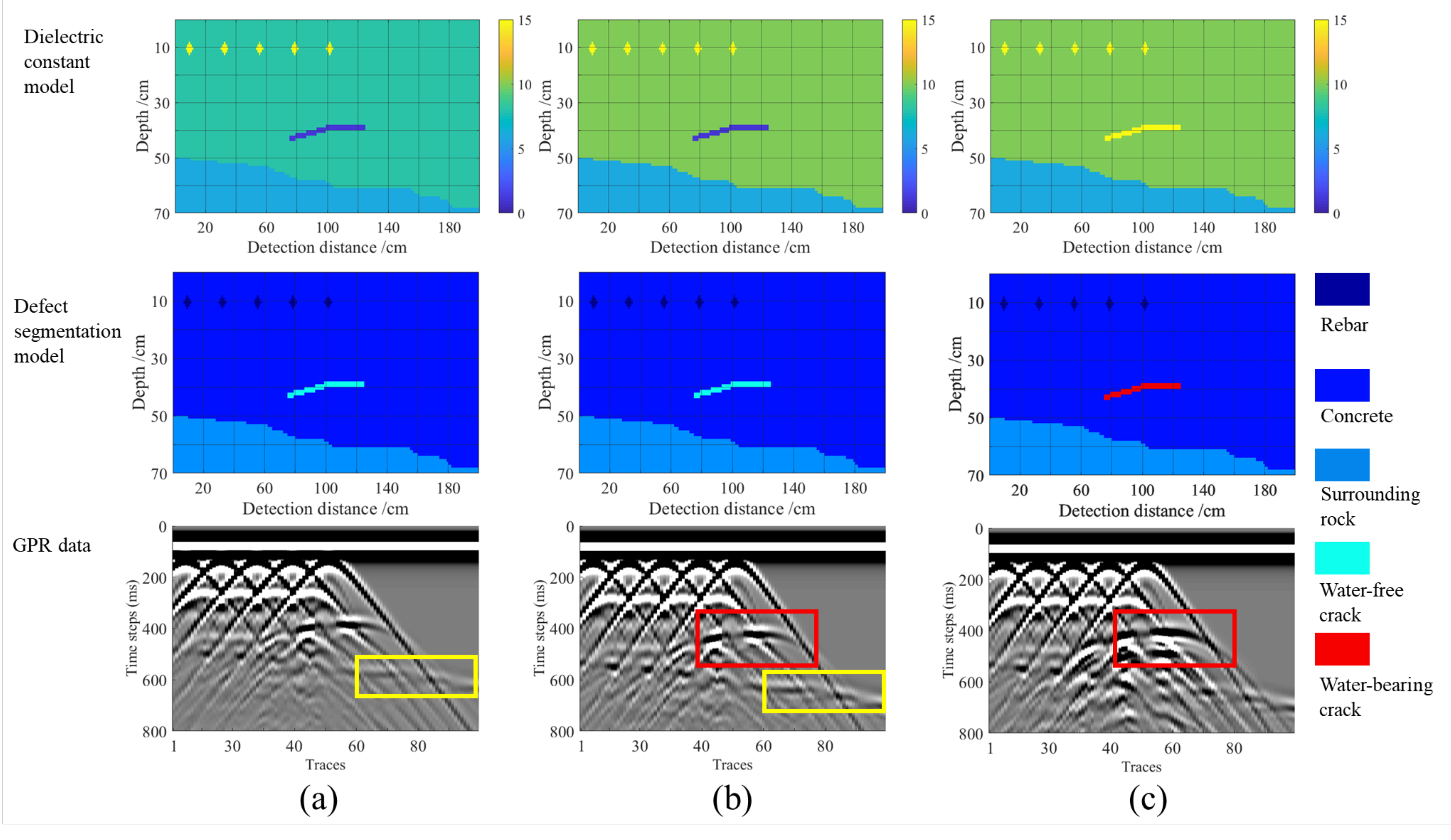}
	\caption{Effect of dielectric constant on GPR signals.}
	\label{fig:4}
\end{figure*}

 The method of defect segmentation is similar to the semantic segmentation often handled in CNNs, but it is also more complicated. These are the following features:
 \begin{enumerate}[(1)]
 \item Firstly, as analyzed by Liu et al.\upcite{liu2019gprinvnet}, GPR data and dielectric constant models differ in shape, value, and distribution. GPR data is a time series measured at different positions in the horizontal direction, and its size is $time steps \times width$, and our target model is a spatial structure with a size of $depth \times width$. There are differences in image processing of detection data between time series and space series. For GPR signals corresponding to different defects, their hyperbola shapes are more similar than natural images. Among them, the shapes of voids and the separations are similar, and their difference lies in their location. The location of the defect needs to be considered, and their differences and features extracted from similar and complex data. In addition, for defects and materials of the same shape, the location and polarity of the reflected signal is affected by different dielectric constants and whether they contain water or not. As shown in Fig.~\ref{fig:4} (a) and (b), the dielectric constants of concrete in the two models are different, which leads to differences in their data, as shown in the yellow box, but the fault segmentation results are the same. Comparing Fig.~\ref{fig:4} (b) and (c), these show how the water content of the defect affects the signal polarity. This considers not only the shape information but also the amplitude of the signal. That is, both shape and value affect the model. The CNN needs to consider both numerical information and shape characteristics. Given that our present problem is that as a classification task, we suggest that a properly selected CNN is more suitable for our defect segmentation than GPRInvNet which was focused on inversion.

\item Normal data and data representing defects are imbalanced in size. As shown in Fig.~\ref{fig:4}(b), the size of the rebars and cracks are small, but they have a great impact on the GPR data. This size imbalance creates a particular problem for the prediction task, that is, defect segmentation. This is especially true for cracks and rebars, as they are smaller in size and require more accurate segmentation effects. To obtain higher resolution prediction results, an effective loss function must be designed and selected.
\item To ensure that all possibilities are included, and to prevent overfitting, a large amount of data and effective measurements are needed for training, and the computational efficiency of the network needs to be taken into account.

\end{enumerate}

Based on the above analysis, we compared Segnet and U-net. It proved that Segnet has the advantages of a simple network structure, fewer parameters, and superior quality results. In addition, Segnet uses max location in upsampling to provide effective information during decoding, which improves the provision of structural information and improved high-frequency data correspondence.

\subsection{Segnet}
\begin{figure*}[hb]
	\centering
	\includegraphics[width=1\linewidth]{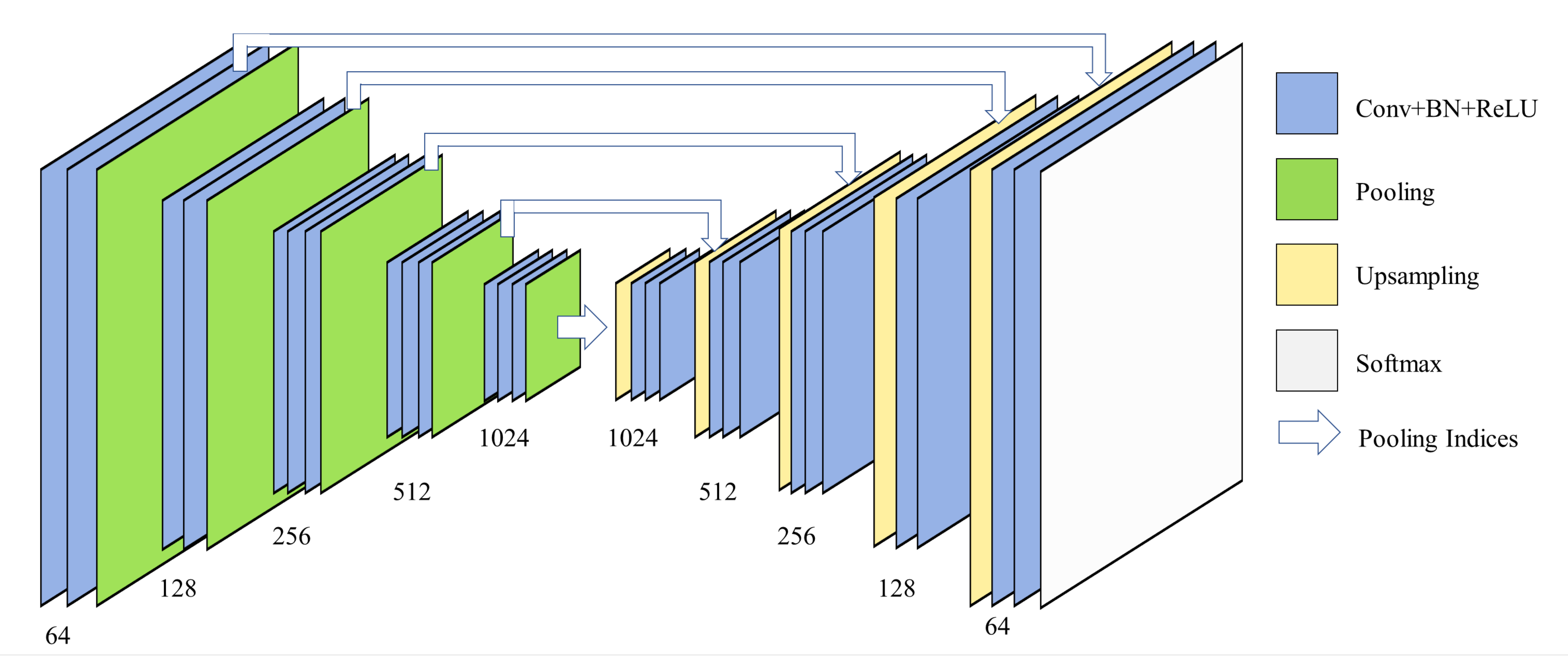}
	\caption{The network structure of Segnet.}
	\label{fig:segnet}
\end{figure*}

The overall structure and idea of Segnet is to use high-dimensional compression data, through convolution and pooling, to obtain high-dimensional features of an image and then up-sampling to complete the regression and segmentation of the image. In the network, the size of the convolution kernel is 3 $\times$ 3. To prevent gradient anomalies, Batch Normalization($BN$) is employed and a Rectified Linear Unit ($ReLU$) is used as the activation function in other layers, with the exception of the last layer. For a semantic segmentation problem, the activation function of the last layer is softmax to obtain the probability under each classification to complete the pixel-level segmentation. The innovation of Segnet is that in the decoder process, the low-resolution feature maps are converted to a high-resolution feature map using the upsampling method. This is different from deconvolution in FCN and U-net. Specifically, in the encoder part, features are compressed by pooling, and the index of each pooling is saved, that is, the original maximum position is saved. Then the corresponding pooling index is used in the decoder for non-linear upsampling. In this way, sparse upsampling feature maps can be obtained without learning the weights used in the deconvolution. Badrinarayanan et al.\upcite{badrinarayanan2017segnet} compared Segnet with common CNNs and proved that Segnet is superior to other methods in classification. Since Segnet is used for image semantic segmentation with fewer parameters and better results, our main purpose was to study the defect segmentation of GPR data by Segnet. Its specific structure is as shown in Fig.~\ref{fig:segnet}. Segnet has few parameters and, in addition, it maintains high-frequency information integrity and has the advantage of achieving improved results compared to competing methods. In Section~\ref{section: results in u}, a comparison between Segnet and U-net proves that Segnet is more suitable for the problem of defect segmentation.
 
\subsection{Loss function}
In the task of semantic segmentation, the most commonly used loss function is the cross entropy loss function. The activation function of the classification problem is softmax, and the use of the L2 norm loss would seriously affect gradient calculations and network updates. The cross entropy loss function can alleviate the problem of gradient disappearance by the logarithm. The cross entropy loss function is as follows:
\begin{equation}
{L_{CE}}=-\frac{1}{p} \sum_{i=1}^{p} \log y_{i}^{*},
\end{equation}
where $i$ is the corresponding position of each pixel, and $y_{i}^{*}$ is the predicted probability of the corresponding position and label.

However, although the effect of the cross entropy loss function has been proven in semantic segmentation, the results of the prediction need to be improved. This improvement is especially necessary for smaller objects, because CNNs using the cross entropy loss function often have difficulty predicting them, making it difficult to achieve the requirements for rebars and cracks prediction in our method. Although rebars and cracks are obviously reflected in the input data, the rebars and cracks are usually very small and require high-resolution processing to be effectively classified in the segmentation. To solve this problem, we applied the Lovász softmax loss function in our CNN. The Lovász softmax loss function was proposed by Berman in 2018 \upcite{berman2018lovasz} to optimize Mean Intersection over Union (MIoU), and its superiority on small objects is proven.
\begin{equation}
L_{L Z S}=\frac{1}{|N|} \sum_{c=1}^{N} \overline{\Delta J c}(m(c)),
\end{equation}
where $\overline{\Delta J c}$ is the Lovász extension to ${\Delta J c}$, approximation to the Jaccard index of class $c$, $N$ is the number of material and defect classes, which is nine in this paper, and $m(c)$ is a vector of pixel errors for class $c$.

Eerapu et al.\upcite{eerapu2019dense} combined the cross entropy and the Lovász softmax loss function to achieve better MIoU. In our work, we also use the composite loss function:
\begin{equation}
L_{sum}={L_{CE}}+L_{L Z S}.
\end{equation}

The results in Section~\ref{section: results in u} show that the addition of the Lovász softmax loss improved the quality of the results, especially for cracks.

\subsection{Network hyperparameters}

Suitable network hyperparameters was beneficial to the training results of the network. PyTorch implemented both CNNs and Adam optimizers in this study. The batch size of the Adam optimizer was set to 24, and the initial learning rate was $5\times10^{-5}$. The network was trained for a total of 100 epochs to obtain sufficient parameter updates.
 
 Considering the similarity of GPR data of different materials and defects, it is easy to overfit the network which seriously affects the generalization ability of the network and even causes unreasonable network parameters. Therefore, effectively avoiding the overfitting problem is an important task. For this, we applied both dropout and weight decay. Dropout was proposed by Hinton\upcite{hinton2012improving} in 2012 and is proven to effectively reduce overfitting to a specific feature by randomly discarding a few per centage points of the features. The weight decay\upcite{krogh1992simple} is used to add a $L_2$ regularization after the loss function, thereby reducing the complexity of the network coefficients, and improving the effectiveness of data fitting. In this paper, by comparison, the dropout probability was set to 20 $\%$ and the weight decay coefficient was $1\times 10^{-4}$, which can effectively alleviate the overfitting of the CNN.

\section{Results and discussion}

\label{section: results}
\begin{figure*}[hb]
	\centering
	\subfigure[]{
		\label{fig:1train}
		\includegraphics[width=0.45\textwidth]{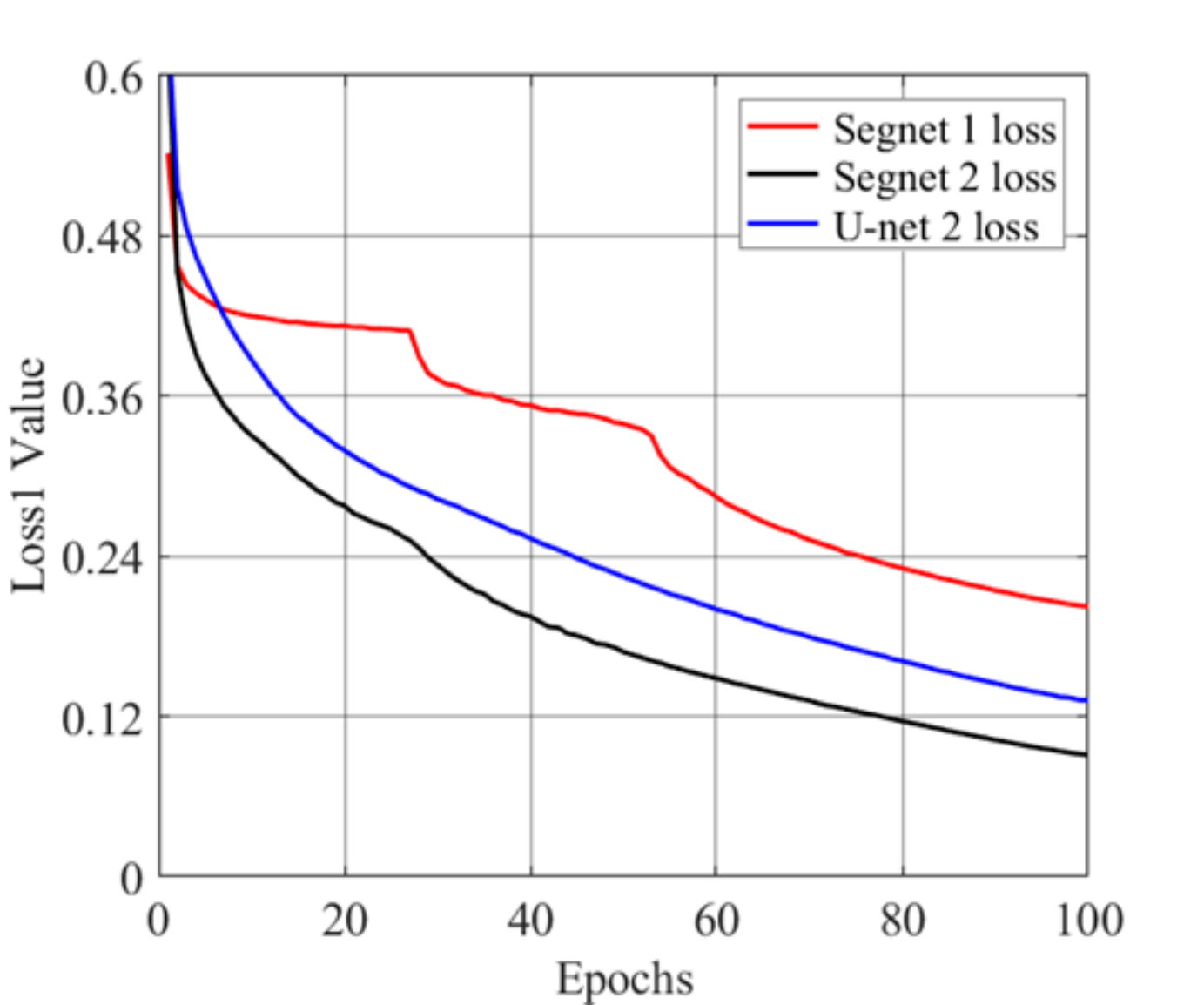}}
	\subfigure[]{
		\label{fig:2train}
		\includegraphics[width=0.45\textwidth]{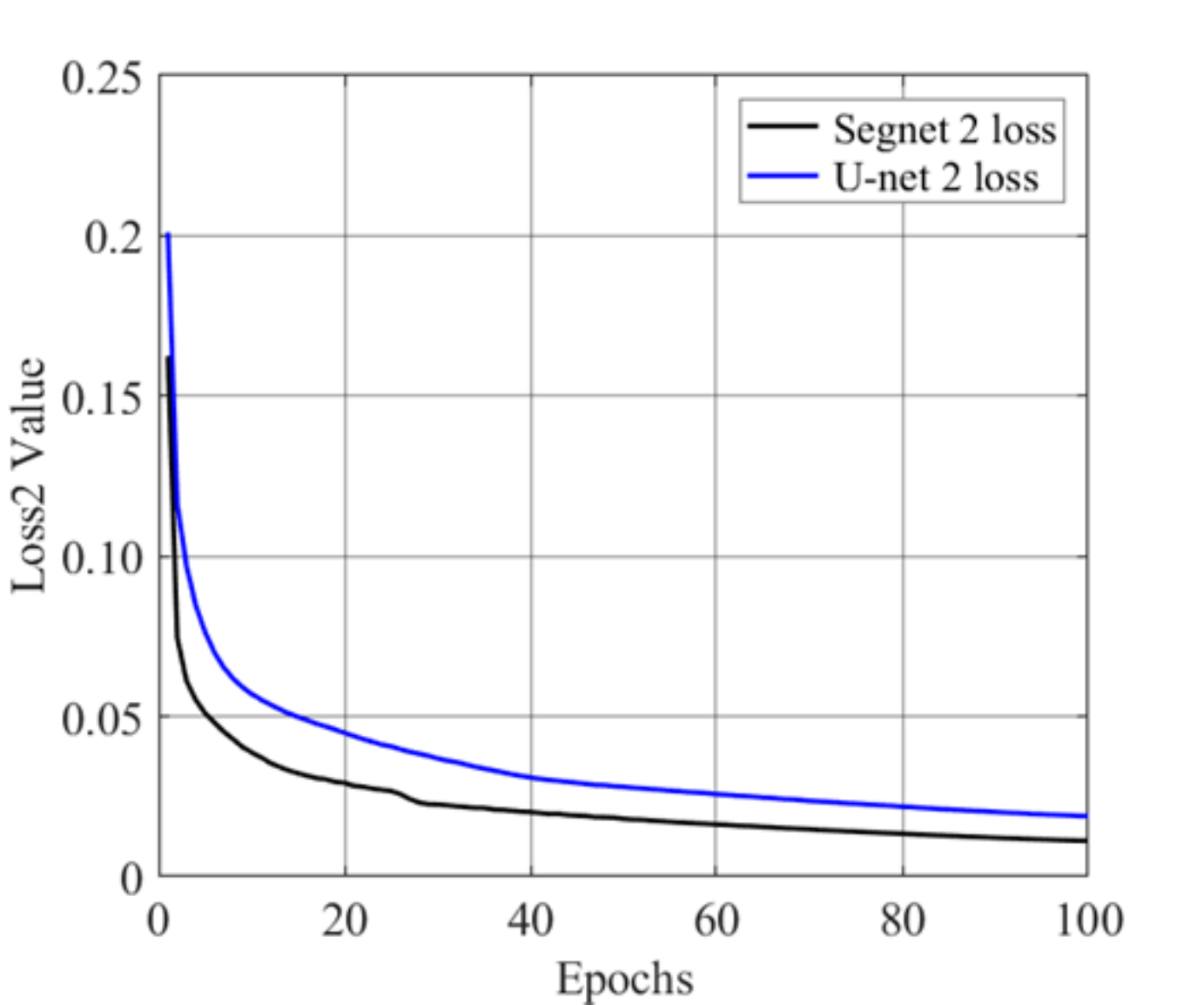}}
	\subfigure[]{
		\label{fig:1valid}
		\includegraphics[width=0.45\textwidth]{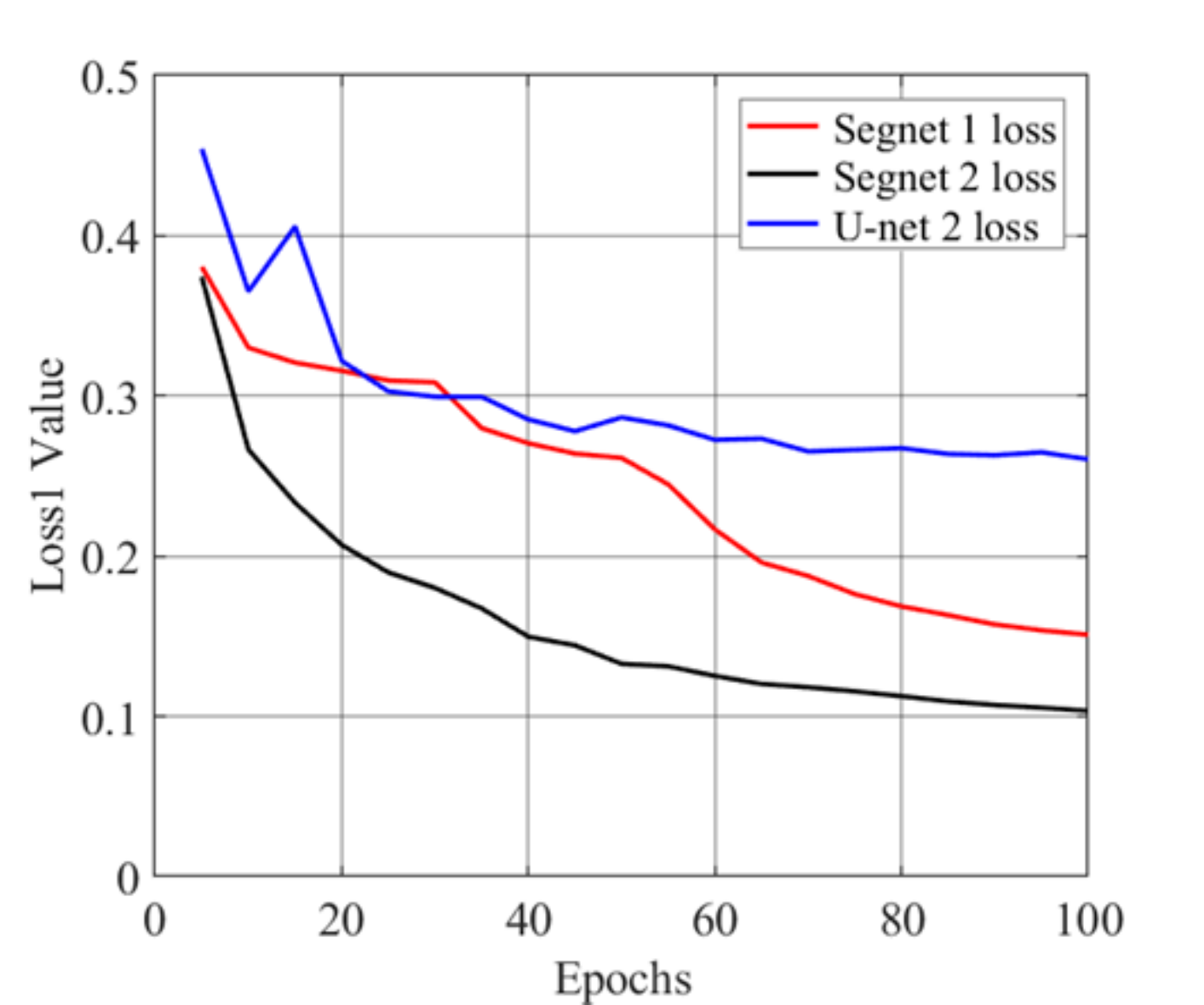}}
	\subfigure[]{
		\label{fig:2valid}
		\includegraphics[width=0.45\textwidth]{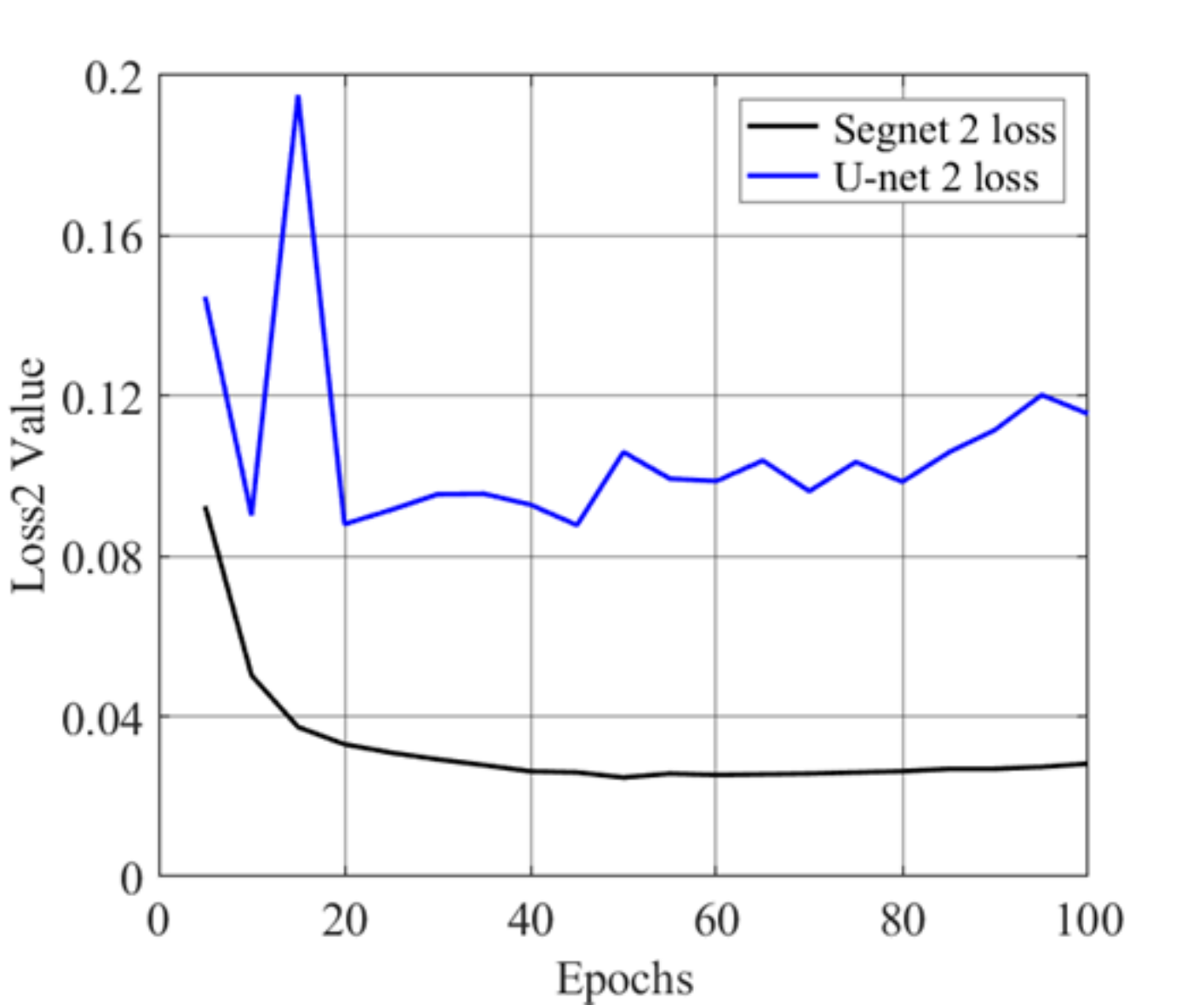}}
	\caption{Loss curves of the three CNN methods. (a) and (b) are the curves of the cross entropy loss function (Loss 1) and the Lovász softmax loss function (Loss 2) on the training set with epoch, respectively. (c) and (d) are the curves of the cross entropy loss function (Loss 1) and the Lovász softmax loss function (Loss 2) on the validation set with epoch, respectively. In these graphs the red, black and blue lines represent Segnet (1 loss), Segnet(2 loss), U-net(2 loss), respectively.}
	\label{fig: loss_curve}
\end{figure*}

To effectively train the network, we divided the data set in Section~\ref{section:data set} into the training, validation, and test set with a ratio of 10:1:1, which were used to train CNNs, verify the ability of the CNNs to save the optimal network parameters, and test the impact of the final network, respectively. Considering the different sizes of input data and output data, we used bicubic interpolation to reshape the input data of $800 \times 99$ into $256 \times 128$, and the output of Segnet was $128 \times 256$ and sized to $90 \times 220$. We trained three CNN-based networks: Segnet using the cross-entry loss function, Segnet using the cross-entry and the Lovász maxsoft loss functions, and U-net using the cross-entry and the Lovász maxsoft loss functions, to compare their effects in the defect segmentation task. To facilitate reference to these three methods, they are named Segnet (1 loss), Segnet(2 loss), and U-net(2 loss), respectively. The loss function curve in the training and validation set is shown in Fig.~\ref{fig: loss_curve}. It can be seen that the result of Segnet(2 loss) is far superior to the comparative methods. As shown in Fig.~\ref{fig: loss_curve}(d), U-net(2 loss) has an overfitting on the validation set, while Segnet(2 loss) does not. In order to quantitatively evaluate the performance of the results of different CNNs and the results of dissimilar materials, a series of indicators were used, such as MPA, MIoU, Precision, and Recall.

\subsection{Metrics}
For a large amount of test set data, it is inconvenient to show each result. Effective evaluation parameters should be used for the statistics of all results, so that the results of different methods and the impact of dissimilar materials can be analyzed. Firstly, for pixel-level classification problems, we used the evaluation metrics commonly employed in semantic segmentation. Mean Pixel Accuracy (MPA), Mean Intersection over Union (MIoU), Frequency Weighted Intersection over Union (FWIoU) were used to evaluate the prediction effect of each model. Secondly, to analyze the effect of our method on each type of material, we also used the Precision, Recall and F-Measure to compare the effects under different types. Moreover, the four types of results with defects were analyzed and the prediction effect of our method in different environments is discussed below.

For MPA, the proportion of correctly classified pixels in each classification was calculated separately, and the average value of all categories was calculated to verify the correctness of the classification. MIoU is the most commonly used standard in classification and semantic segmentation. It calculates the ratio of the intersection and union of the true and predicted values for each category and calculates the average. FWIoU is an improvement on MIoU, which sets weights for each class based on how often it appears. Considering that these three evaluation metrics are very commonly used in semantic segmentation\upcite{garcia2017review}, we used them to evaluate the similarity of each prediction result and ground truth. Precision and recall are, respectively, the correct probability of classifying pixels, and for the probability of correct classification of ground truth, while F-measure is a hybrid metric which is the harmonic mean of precision and recall\upcite{martin2004learning}. They are used to evaluate the results of each type of material and defect prediction.

\subsection{Comparison of results with U-net and different loss functions}
\label{section: results in u}

\begin{figure*}[hb]
	\centering
	\includegraphics[width=0.8\linewidth]{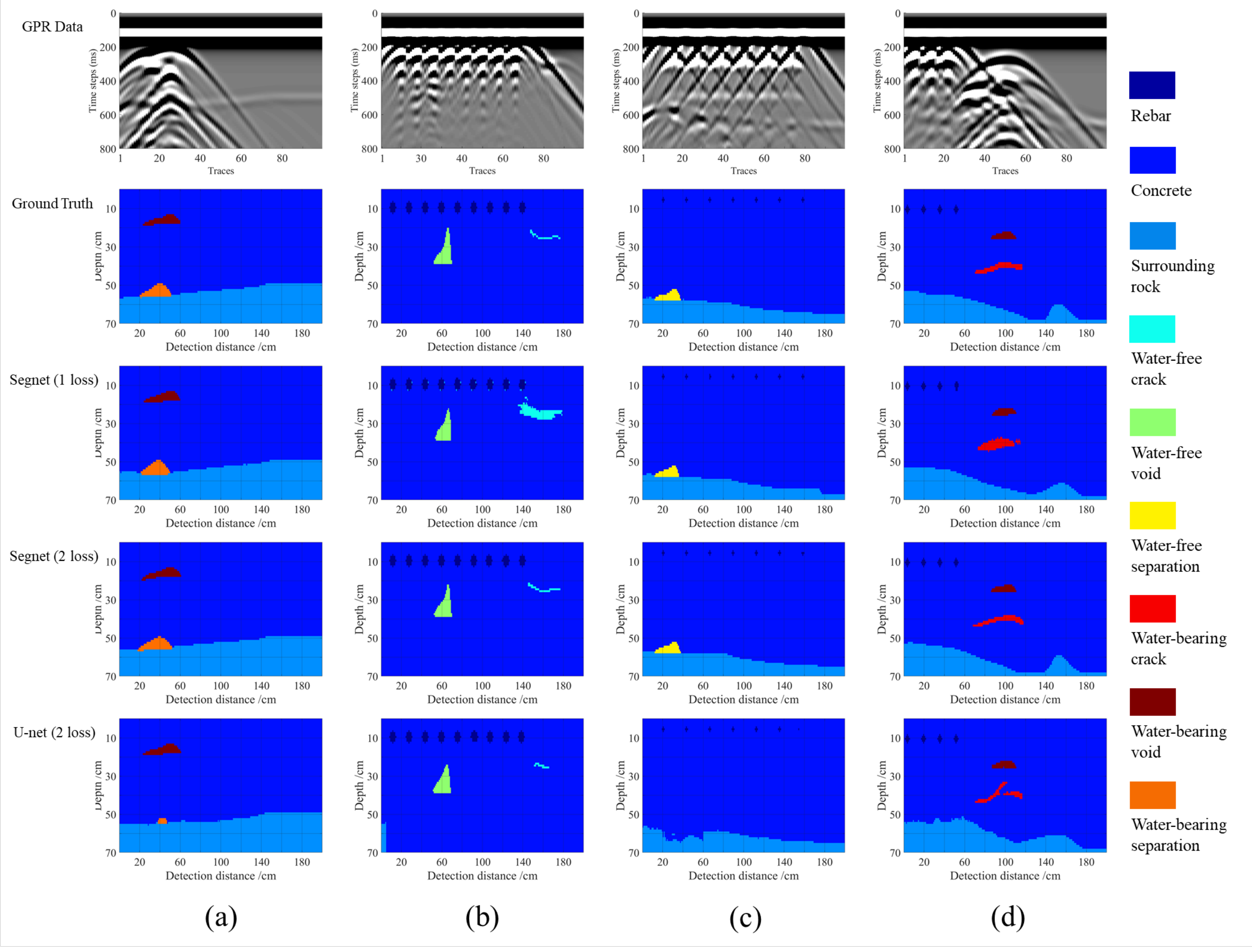}
	\caption{Prediction results of three methods on test set data. It contains four sets of data, ground truth, and prediction results of Segnet(1 loss), Segnet(2 loss), and U-net(2 loss).}
	\label{figure:result_sum}
\end{figure*}

The comparison above of the loss functions has shown the effectiveness of Segnet(2 loss) in tunnel lining defect segmentation. Similarly, we also compares the performance of the three methods on the test set in Table ~\ref{Tab:02}, which also demonstrates the performance of Segnet(2 loss). Specifically, we selected four typical model-data pairs, as shown in Fig. \ref{figure:result_sum}. In conjunction with Table ~\ref{Tab:03}, the effects of each material and defect under each method were analyzed.

Firstly, Segnet(1 loss) achieved acceptable results and had very accurate predictions for rebars, voids, and separations. However, it performed poorly for cracks and had the lowest precision. The recognition of cracks requires high resolution, which is more difficult for segmentation problems. As shown in Fig.\ref{figure:result_sum} (b) and (d), it can be seen that this method had a lower prediction accuracy for thinner defects such as cracks. The cross-entropy loss function focuses on the probability of each pixel but is limited to the overall effect, which results in a lower resolution of the result, making it difficult to identify small defects.

Secondly, Segnet(2 loss) performed optimally in all materials and defects. It accurately predicted the location and classification of cracks, voids, and separations, and was good for complex data. Good results are obtained for smaller defects, such as cracks and rebars. In general, most of the prediction results are accurate. The presence of reinforcing bars may cause some results to be incorrect, but the probability of errors is very low. This result proves the effectiveness of the Lovász Softmax loss and our method.

In addition, the effect of U-net(2 loss) was poorer than that of Segnet(1 loss). When multiple defects occurred at different depths in the same location, GPR data was more complicated. The U-net(2 loss) made it difficult to effectively classify the defects, especially separations and cracks with water, as shown in Fig.\ref{figure:result_sum}. The lack of accurate defect prediction would seriously affect the judgment of the integrity of the tunnel lining. Through analysis, we believe that it is the network structure of U-net that has led to poor results. The characteristic of U-net is that it saves the features of the encoded segments and uses them as feature maps when decoding, which is effective for the task of one-to-one correspondence in image semantic segmentation. However, for our method, GPR data and defect segmentation were not completely corresponding, which caused U-net to introduce incorrect information and obtain poor results.
 
\begin{table}[]
\footnotesize
\centering
\caption{Results of different methods}
\label{Tab:02}
\begin{tabular}{ccccccc}
Metrics & Segnet(1 loss) & Segnet(2 loss)  & U-net(2 loss)  \\ \hline
MPA     & 0.91  & 0.93 & 0.79   \\ \hline
MIoU    & 0.83  & 0.90 & 0.75       \\ \hline
FWIoU   & 0.99  & 0.98 & 0.96  \\ \hline  
\end{tabular}
\end{table}

\begin{table}[]
\centering
\caption{Results of different methods and different categories}
\label{Tab:03}
\resizebox{\textwidth}{!}
{
\begin{tabular}{cccllllllllll}
\multicolumn{1}{l}{\multirow{2}{*}{Metrics}} & \multirow{2}{*}{Methods} & \multicolumn{3}{c}{Linging Matrials} &  & \multicolumn{3}{c}{Defects} &  & \multicolumn{3}{l}{Water-bearing Defects} \\ \cline{3-5} \cline{7-9} \cline{11-13}
\multicolumn{1}{l}{}                         &                          & Rebar     & Concrete    & Rock      &  & Crack   & Void    & Separation  &  & Crack        & Void         & Separation      \\ \hline
\multirow{3}{*}{Precision}                   & Segnet(2 loss)            & 0.9664    & 0.9968      & 0.9765    &  & 0.8157  & 0.9008     & 0.9111  &  & 0.8833       & 0.8278       & 0.8798      \\
                                             & Segnet(1 loss)            & 0.9312    & 0.9961      & 0.9698    &  & 0.3108  & 0.8674  & 0.9019  &  & 0.8754       & 0.8441       & 0.8754      \\
                                             & U-net(2 loss)             & 0.9543    & 0.9875      & 0.9242    &  & 0.6324  & 0.8109  & 0.2918  &  & 0.2619       & 0.7807       & 0.2619      \\ \hline
\multirow{3}{*}{Recall}                      & Segnet(2 loss)            & 0.9541    & 0.9969      & 0.9777    &  & 0.8039  & 0.8869  & 0.8960  &  & 0.8740       & 0.8303       & 0.8715      \\
                                             & Segnet(1 loss)            & 0.9201    & 0.9928      & 0.9747    &  & 0.7889  & 0.8635  & 0.8923  &  & 0.8703       & 0.8379       & 0.8703      \\
                                             & U-net(2 loss)             & 0.9449    & 0.9861      & 0.9121    &  & 0.6495  & 0.8001  & 0.193   &  & 0.1729       & 0.7629       & 0.1729      \\ \hline
\multirow{3}{*}{F-measure}                   & Segnet(2 loss)            & 0.9602    & 0.9968      & 0.9771    &  & 0.8098  & 0.8938  & 0.9035   &  & 0.8786       & 0.8290       & 0.8756      \\
                                             & Segnet(1 loss)            & 0.9256    & 0.9945      & 0.9723    &  & 0.4459  & 0.8654  & 0.8971  &  & 0.8728       & 0.841        & 0.8728      \\
                                             & U-net(2 loss)             & 0.9495    & 0.9868      & 0.9181    &  & 0.6408  & 0.8054  & 0.2324  &  & 0.2083       & 0.7717       & 0.2083      \\ \hline
\end{tabular}
}
\end{table}

\subsection{Results in dissimilar materials }
\label{section: results in dm}

Through the above analysis, we demonstrated the effectiveness of Segnet and the Lovász softmax loss on defect segmentation and explained the unsuitability of U-net. Segnet also has different performance effects for different types of materials and defects. We divided the defect models into four categories and analyzed them in turn.
\begin{enumerate}[(1)]

\item Water-free defects in the tunnel lining without rebars
For defects in the tunnel lining without rebars, because the model is simple, our method obtained accurate classification, location and morphology in various defects, as shown in Fig.~\ref{fig:7} and Table ~\ref{Tab:04}. Correct predictions at the pixel level are achieved on all models.

 \begin{figure*}[hb]
	\centering
	\includegraphics[width=0.8\linewidth]{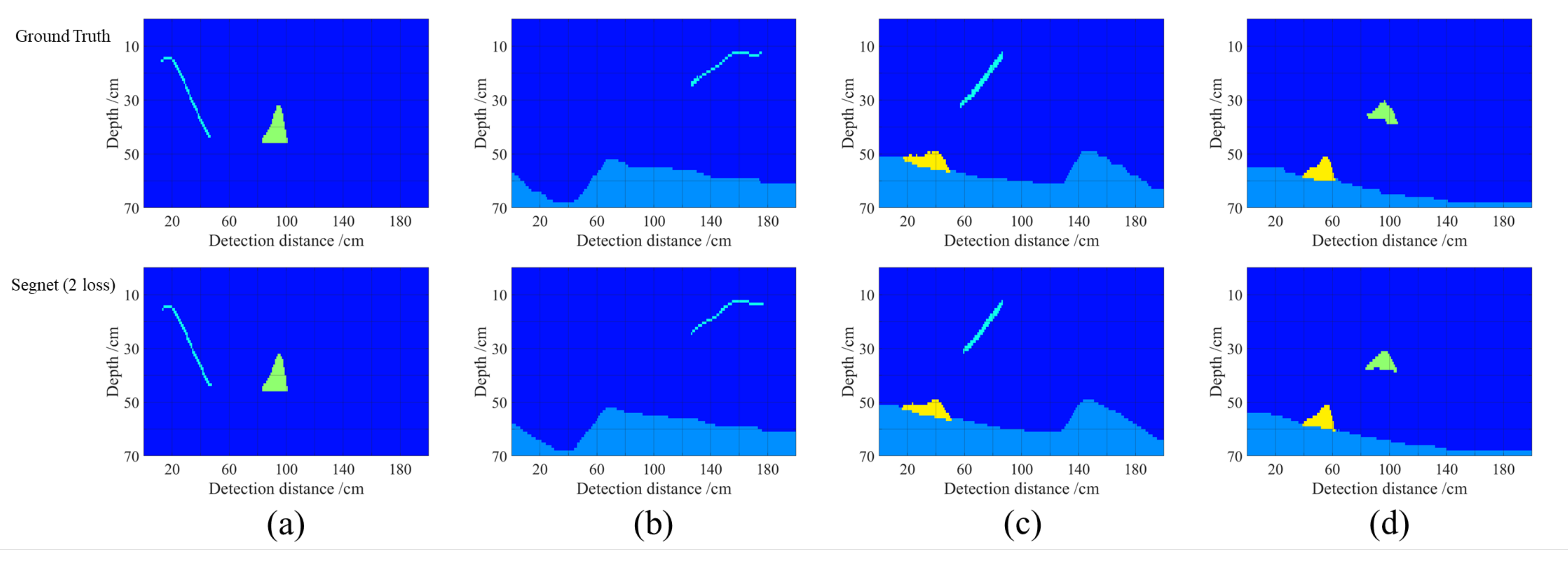}
	\caption{Water-free defects in the tunnel lining without rebar model prediction results. (a) represents the water-free crack and void model without surrounding rock. (b) represents the water-free crack model. (c) represents the water-free crack and separation model. (d) represents the water-free void and separation model. Each color represents the same material as in Fig.~\ref{figure:result_sum}.}
	\label{fig:7}
\end{figure*}

\begin{table}[]
\centering
\caption{Results of defection detection without rebars in different types of models}
\label{Tab:04}
\footnotesize
\resizebox{\textwidth}{!}
{
\begin{tabular}{cccccccc}
Class & Defects & Crack & Void & Separation & Crack\&Void & Crack\&Separation & Void\&Separation \\ \hline
\multirow{3}{*}{Water-free}& MPA     & 0.96  & 0.98 & 0.97 & 0.91     & 0.95       & 0.95         \\ \cline{2-8}
& MIoU   & 0.93  & 0.96 & 0.95 & 0.86      & 0.91     & 0.92         \\ \cline{2-8}
& FWIoU   & 0.99  & 0.99 & 0.99 & 0.99        & 0.99          & 0.99      \\ \hline  
\multirow{3}{*}{Water-bearing}& MPA      & 0.96  & 0.97 & 0.97 & 0.90        & 0.93          & 0.94         \\ \cline{2-8}
& MIoU    & 0.93  & 0.96 & 0.95 & 0.84        & 0.88          & 0.91         \\ \cline{2-8}
& FWIoU  & 0.99  & 0.99 & 0.99 & 0.98        & 0.98          & 0.99        \\ \hline
\end{tabular}
}
\end{table}

\item Water-bearing defects in the tunnel lining without rebars
Similarly, the water-bearing defect was relatively simple, so the overall effect was satisfactory. It can be seen in Table~\ref{Tab:04} that crac detection performed less well than other defects especially the model of cracks and voids, which shows that water-bearing defects had an impact on the results. As shown in Fig.~\ref{fig:8} (c), especially when the void and crack are in the same horizontal position, the upper defect affects the data below, making it difficult to obtain an accurate shape, but the classification of the defects was accurate.

 \begin{figure*}[hb]
	\centering
	\includegraphics[width=0.8\linewidth]{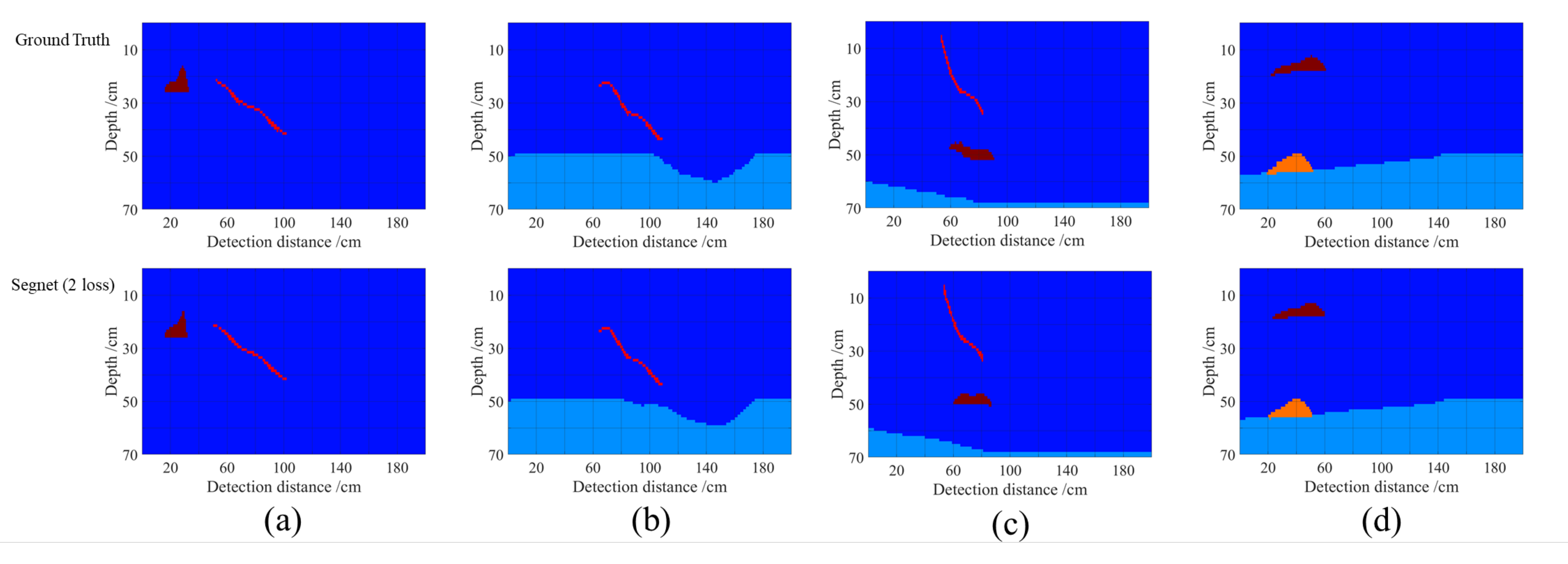}
	\caption{Water-bearing defects in the tunnel lining without rebar model prediction results. (a) represents the water-bearing crack and void model without surrounding rock. (b) represents the water-bearing cracks model. (c) represents the water-bearing crack and separation model. (d) represents the water-bearing void and separation model. Each color represents the same material as in Fig.~\ref{figure:result_sum}.}
	\label{fig:8}
\end{figure*}


\item Defect in the tunnel lining with rebars
The most challenging aspect of this method was that the rebars in the lining would seriously affect the internal defect signals acquisition. A row of rebars with a small diameter are reflected in GPR data as multiple parallel hyperbolae. They intersect each other, shielding the effective information below. In our results, the effect of rebars on the defect, especially cracks, was severe. As shown in Table ~\ref{Tab:06}, the models of cracks and voids under rebars, whether they contain water or not, have a MIoU value below 0.8, which was very rare in our results otherwise. As shown in Fig.~\ref{fig:9}(c), the length of the crack was also incorrectly predicted, and the interface of the rock was also inaccurate in Fig.~\ref{fig:9} (d). For our judgment of defects, correct classification and accurate positioning can meet most of our requirements, so the above problems have little impact on us.

\begin{figure*}[hb]
	\centering
	\includegraphics[width=0.8\linewidth]{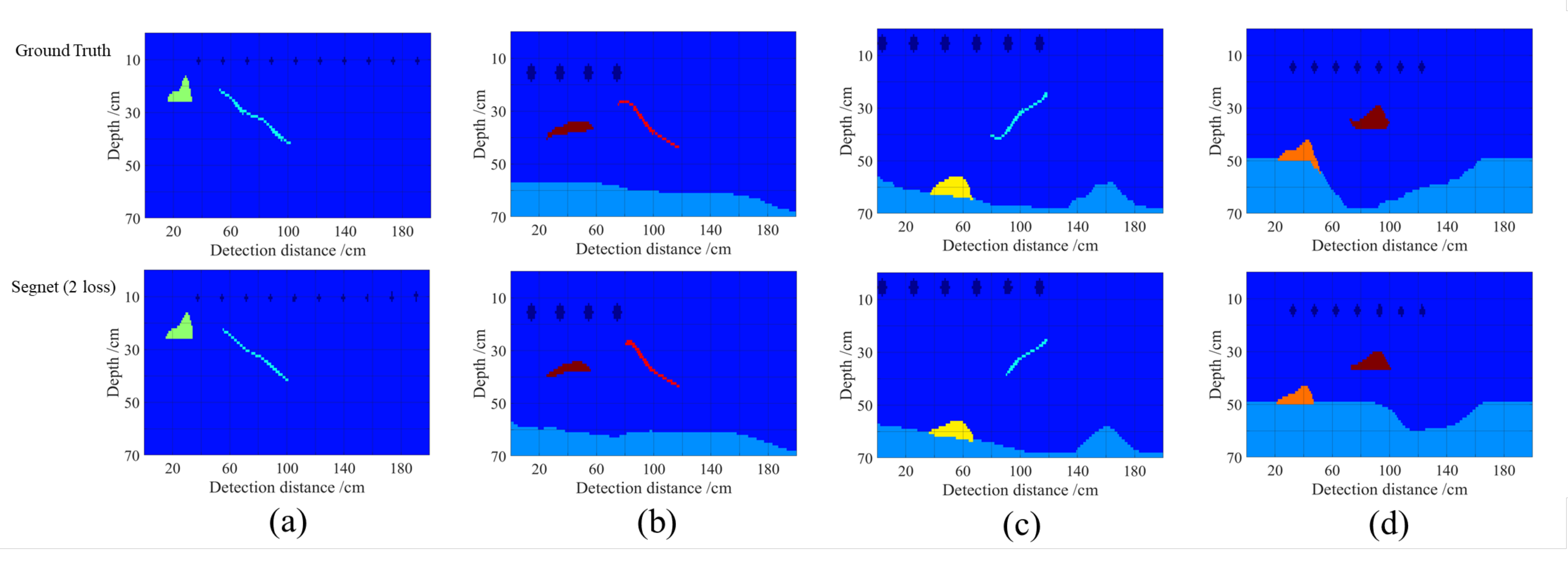}
	\caption{Water-bearing defects in the tunnel lining without rebar model prediction results. (a) represents the water-free crack and void model with rebars. (b) represents the water-bearing crack and void model with rebars. (c) represents the water-free crack and separation model with rebars. (d) represents the water-bearing void and separation model with rebars. Each color represents the same material as in Fig.~\ref{figure:result_sum}.}
	\label{fig:9}
\end{figure*}

\begin{table}[]
\centering
\caption{Results of defection detection without rebars in different types of model}
\label{Tab:06}
\footnotesize
\resizebox{\textwidth}{!}
{
\begin{tabular}{cccccccc}
Class & Defects & Crack & Void & Separation & Crack\&Void & Crack\&Separation & Void\&Separation \\ \hline
\multirow{3}{*}{Water-free}& MPA     & 0.91  & 0.95 & 0.96 & 0.84        & 0.89          & 0.91         \\  \cline{2-8}
& MIoU    & 0.86  & 0.91 & 0.94 & 0.78        & 0.84          & 0.87         \\  \cline{2-8}
& FWIoU   & 0.99  & 0.98 & 0.99 & 0.98        & 0.98          & 0.98        \\ \hline
\multirow{3}{*}{Water-bearing}& MPA     & 0.91  & 0.94 & 0.96 & 0.83        & 0.89          & 0.90         \\ \cline{2-8}
& MIoU    & 0.86  & 0.90 & 0.94 & 0.76        & 0.83          & 0.85         \\  \cline{2-8}
& FWIoU   & 0.98  & 0.98 & 0.99 & 0.97        & 0.97          & 0.98        \\ \hline
\end{tabular}
}
\end{table}

\end{enumerate}

In general, our Segnet with two loss functions achieved excellent defect segmentation results, which was far better than Segnet with one loss function and U-net with two loss functions. In linings containing cracks and rebars, accurate classification and positioning could obtained, which also reflects the high accuracy of our method. Although there were some problems for defect detection under rebars, in most models the correct classification could still be obtained, which provides an effective reference for post-processing. Due to the complexity of the data, some errors are inevitable we believe.

\section{Experiment on model testing }
\label{section: model}
As a result of our network design and training, we obtained a CNN model which achieved excellent results on synthetic test data. Real data is more complicated than synthetic data, and noise and other disturbances seriously affect the quality of the collected data. In geophysics, because the detection area is usually unknown, many studies use physical model tests to verify the viability and applicability of theoretical methods\upcite{xie2013gpr, sivaji2002physical, liu2020rapid}. To obtain GPR data from a known internal structure and verify the effect in a real environment, we built a test model to simulate the internal defects of a real tunnel lining. Real GPR data was collected, analyzed, and processed. At the same time, the CNN was fine-tuned to fit the real data. Finally, we used the CNN to segment the defects inside the tunnel lining with real data.

\subsection{Model test building}
\begin{figure*}[hb]
	\centering
		\label{fig:model1}
		\includegraphics[width=0.8\textwidth]{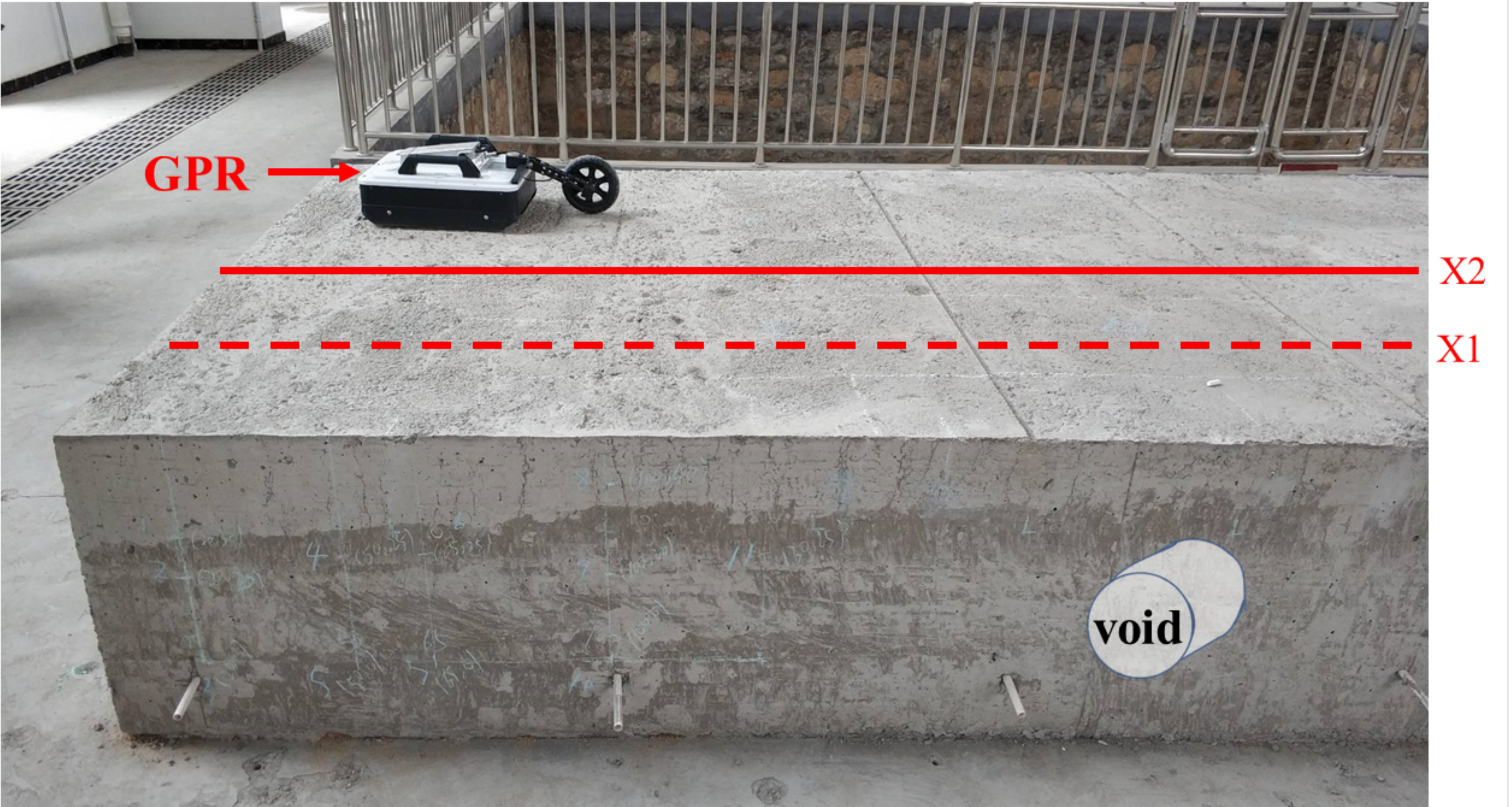}
	\caption{The model we built and GPR for detection.}
	\label{fig:model testing}
\end{figure*}

We designed a model test using a concrete testbed with a size of $4.4m \times 2m \times 0.7m$ to simulate the internal structure of a tunnel lining, as shown in Fig.~\ref{fig:model testing}. In the model, we used the materials employed in the lining of an actual tunnel. To simulate a water-bearing void, we used a PVC pipe with a length of 400 $mm$ and a diameter of 120 $mm$ to construct the separation. PVC pipes were filled with water, sealed at both ends and placed in the concrete to simulate a water-bearing void in the lining. As a result of this construction method, we knew the exact materials used and their shape within the model. On the model, we set two sidelines with a distance of $0.5m$ and a length of $4.4m$ to obtain reflection information of the embedded defects below the model, of which there is no defect below the X2 sideline. Through the X2 sideline, we could get enough background data under the current model for further experimental analysis and processing. The real GPR data was collected by Impulse Radar 600MHz equipment, with 512 sampling point. The mode of the GPR was set to `Wheel', and the distance of the traces was chosen as 0.02$m$. The GPR data was collected and transmitted to the computer via Wi-Fi, and the data was initially processed.

\subsection{Data processing}
For the actual measured data, referring to previous research\upcite{liu2019gprinvnet}, we preprocessed the real GPR data, including static correction, and removed the direct component, background signal, and bandpass filtering. In this way, more effective GPR data could be acquired, which improved the effectiveness of our method. To be more suitable for the real data, we used actual data from measurements of the plain concrete without defects, and randomly added it to the synthetic data as background noise. Because the size of the measured data and the synthesized data were different, we adjusted the measured data by the bicubic difference method and obtained hundreds of sets of background noise. These actual noise data sets were normalized, fixed to a reasonable range, and randomly added to the synthesized data. The updated synthetic data was used to train and fine-tune the CNN parameters for 40 epochs. In this way, the non-uniformity of the actual medium and the interference noise collected were considered, and the applicability of the CNN was further improved. This also provided a useful reference for real data processing under different operating conditions and environments in the CNNs method.

\subsection{Result on Real Data}
\begin{figure*}[hb]
	\centering
	\includegraphics[width=0.9\linewidth]{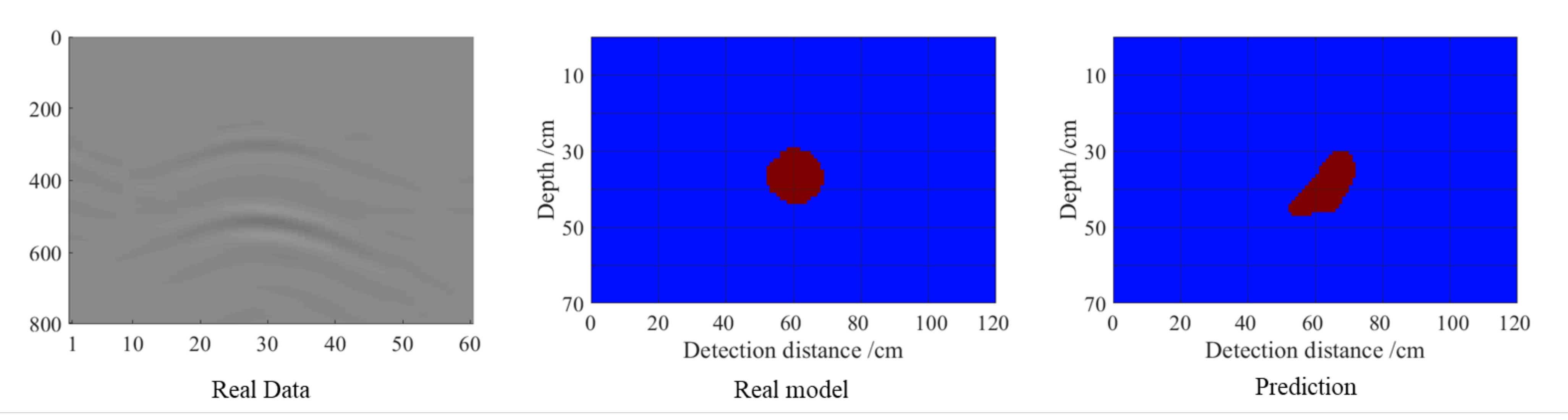}
	\caption{Results of the real GPR data. These show measured GPR data, corresponding model, and the model prediction respectively.}
	\label{fig:real}
\end{figure*}

We tested the retrained CNN with the real data on the X1 sidelines. After finishing the retrained CNN and data processing, the internal structure information can be directly obtained, which improves the automation and efficiency of GPR data interpretation. For water-bearing voids, as shown in Fig.~\ref{fig:real}(a), our method correctly predicted the classification and location of the defect, but the shape of the defect was poorly predicted. In the case of a single defect, our method achieves excellent results, which proves the feasibility of our method. The potential of our method for real data was demonstrated through model building and effective data processing. It was useful to fine-tune the network based on the addition of real background and synthetic data, which may be necessary for the process of applying the CNNs method to real data. 

\section{Conclusion}

In this paper, we applied a CNN to create a new method called defect segmentation to resolve the problem of GPR data interpretation and tunnel lining defects detection. The conclusions of this study are as follows:
\begin{enumerate}[(1)]
\item There are many differences between the defect segmentation of GPR data and semantic segmentation of natural images, including the dissimilarity of the signals, the morphological differences between the input and output, and the effect of the values on the results, which makes it difficult to apply CNNs to the task of GPR defect segmentation directly.
\item The characteristics of Segnet make it more suitable for our method than U-net and we have shown that it achieved more accurate results. Almost all models in a synthetic dataset were correctly classified, and an MPA of 93$\%$ and MIoU of 90$\%$ have been achieved with the cross entry and Lovász softmax loss function.
\item Due to the effect of the method greatly improving the segmentation accuracy, the Lovász softmax loss function is worth recommending, especially for crack detection. Segnet combining the cross entropy and the Lovász softmax loss function improved 7$\%$ the MIoU as compared with Segnet, which only uses cross entropy.
\item For the CNN method, the complexity of the data also directly led to the accuracy of the prediction results. Both water-bearing defects and rebars had an impact on the segmentation problem. In particular, the presence of rebars and the response of the GPR signal to these, seriously affected the signals of underlying defects and this caused prediction difficulties. 
\item When applying our proposed CNN to real data, we suggest that the appropriate methodology is to collect background signals of the corresponding environment, combine existing synthetic data sets, and fine-tune the network to achieve better results.
\end{enumerate}

\section*{Acknowledgments}
 
The research was supported by the National Key R \& D Program of China (No. 2018YFC0406904), Joint Program of the National Natural Science Foundation of China (No. U1806226), The key project of National Natural Science
Foundation of China (No. 51739007), and Shandong Provincial Natural Science Foundation (No.ZR2018MEE052).


\bibliography{defect-segmentation}

\end{document}